\pdfoutput=1

\documentclass{article}

\usepackage[final]{nips_2016}

\usepackage[utf8]{inputenc} 
\usepackage[T1]{fontenc}    
\usepackage{url}            
\usepackage{booktabs}       
\usepackage{amsfonts}       
\usepackage{nicefrac}       
\usepackage{microtype}      

\usepackage{amsmath,amsthm,verbatim,amssymb,amsfonts,amscd, graphicx}
\usepackage{graphics}

\usepackage{graphicx}
\usepackage{adjustbox}

\usepackage{color}

\theoremstyle{plain}
\newtheorem{theorem}{Theorem}

\theoremstyle{definition}

\newcommand{\norm}[1]{\left|\left| #1 \right|\right|}

\usepackage{xcolor}
\definecolor{darkgreen}{rgb}{0,0.6,0}
\definecolor{darkred}{rgb}{0.6,0,0}

\newcommand{\mbold}[1]{\mathbb{#1}}

\makeatletter
\@fpsep\textheight
\makeatother

\setcitestyle{square,numbers,comma}

\title{
Survey of Expressivity in Deep Neural Networks
}

\author{
Maithra Raghu \\
Google Brain and Cornell University
\And
Ben Poole \\
Stanford University and Google Brain
\And 
Jon Kleinberg \\
Cornell University
\And
Surya Ganguli \\
Stanford University
\And
Jascha Sohl-Dickstein \\
Google Brain
}

\begin{document}

\maketitle

\begin{abstract}
    We survey results on neural network expressivity described in \cite{raghu2016expressivity}. The paper motivates and develops three natural 
    measures of expressiveness, which all display an exponential dependence on the depth of the network. In fact, all of these measures are related to a fourth quantity, \textit{trajectory length}. This quantity grows exponentially in the depth of the network, and is responsible for the depth sensitivity observed. These results translate to consequences for networks during and after training. They suggest that parameters earlier in a network have greater influence on its expressive power -- in particular, given a layer, its influence on expressivity is determined by the \textit{remaining depth} of the network after that layer. This is verified with experiments on MNIST and CIFAR-10. We also explore the effect of training on the input-output map, and find that it trades off between the stability and expressivity.
\end{abstract}

\section{Motivation and Setting} 
\label{subsec_Express}
In this survey, we summarize results on the {\em expressivity} of deep neural networks from \cite{raghu2016expressivity}. Neural network expressivity looks at how the architecture of the network (width, depth, connectivity) affects the properties of the resulting function. 

Being a fundamental step to better understanding neural networks, there is much prior work in this area. Many of the existing results rely on comparing \textit{achievable functions} of a particular network architecture,  (\cite{hornik1989multilayer,cybenko1989approximation}, \cite{eldan2015power,telgarsky2015representation,martens2013representational,bianchini2014complexity}). While compelling, these results also highlight limitations of much of the existing work on expressivity -- unrealistic assumptions are sometimes made about the architectural shape e.g. exponentially large width, and networks are often compared via their ability to approximate one \textit{specific} function, which, in isolation, cannot result in a more general conclusion.

To overcome this, we start by analyzing expressiveness in a setting which is both more general than one of hardcoded functions, and immediately related to practice -- networks after \textit{random initialization.} Not only does this mean conclusions are independent of specific weight settings, but understanding behavior at random initialization provides a natural baseline to compare to the effects of training and trained networks, which we summarize in Sections \ref{sec_training}, \ref{sec_trained}.

\paragraph{Companion Paper} In a companion paper, \cite{poole2016exponential}, the propagation of \textit{Riemannian curvature} through random networks is studied by developing a mean field theory approach, which quantitatively supports the conjecture that deep networks can disentangle curved manifolds in input space.

\section{Random networks}

The results on networks after random initialization examine the effect of depth and width of a network architecture on its expressive power after random initialization via three natural measures of functional richness, number of transitions, activation patterns, and dichotomies. More precisely, fully connected networks of input dimension $m$, depth $n$ and width $k$ are studied, with weights, bias randomly initialized as $\sim N(0, \sigma_w^2/k), N(0, \sigma_b^2)$.

\subsection{Measures of Expressivity}
In more detail, the measures of expressivity are:

\textbf{Transitions:} Counting neuron transitions is introduced indirectly via linear regions in  \cite{pascanu2013number}, and provides a tractable method to estimate the non-linearity of the computed function. 

\begin{figure}
\centering
\adjincludegraphics[width=0.8\linewidth]{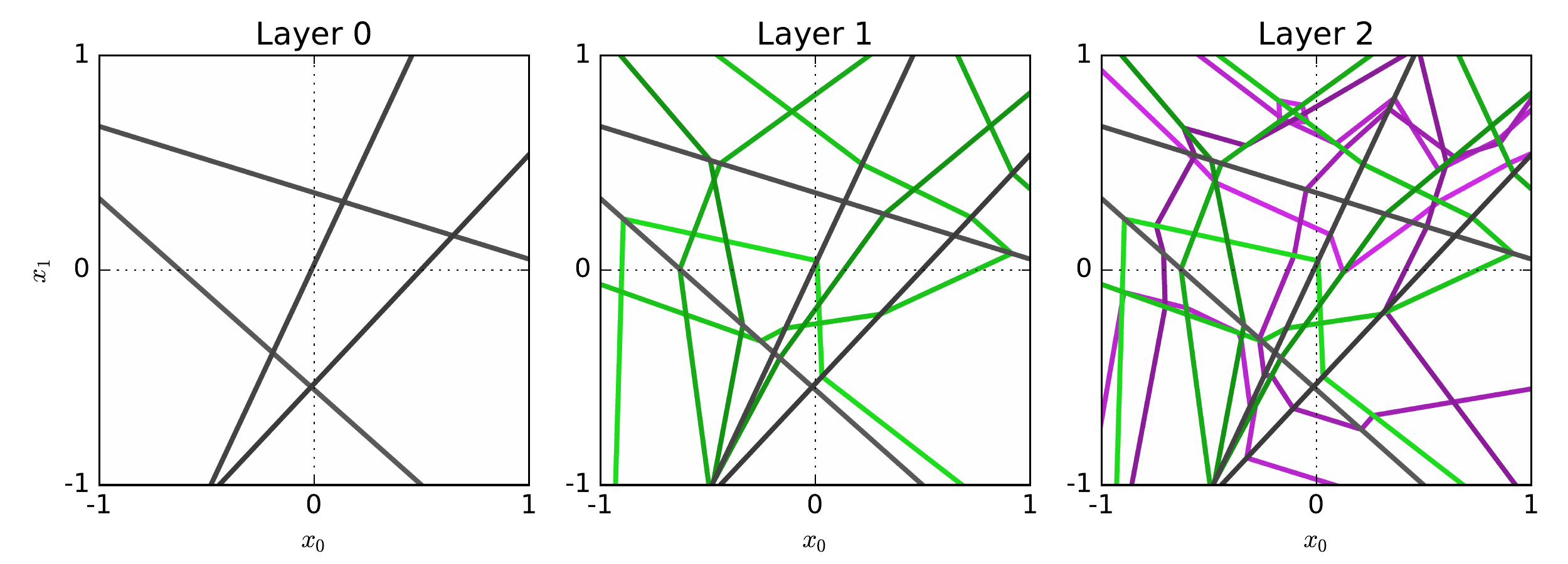}
\caption{
Deep networks with piecewise linear activations subdivide input space into convex polytopes.
Here we plot the boundaries in input space separating unit activation and inactivation for all units in a three layer ReLU network, with four units in each layer. 
The left pane shows activation boundaries (corresponding to a hyperplane arrangement) in gray for the first layer only, partitioning the plane into regions. The center pane shows activation boundaries for the first two layers. Inside \textit{every} first layer region, the second layer activation boundaries form a \textit{different} hyperplane arrangement. The right pane shows activation boundaries for the first three layers, with different hyperplane arrangements inside all first and second layer regions. This final set of convex regions correspond to different activation patterns of the network -- i.e. different linear functions.
\label{fig convex polytope}
}
\end{figure}

\textbf{Activation Patterns:} Transitions of a single neuron can be extended to the outputs of all neurons in all layers, leading to the (global) definition of a network \textit{activation pattern}, also a measure of non-linearity. Network activation patterns directly show how the network partitions input space (into \emph{convex polytopes}), through connections to the theory of \textit{hyperplane arrangements}, Figure \ref{fig convex polytope}.

\textbf{Dichotomies:} The \textit{heterogeneity} of a generic class of functions from a particular architecture is also measured, by counting the number of dichotomies seen for a fixed set of inputs. This measure is `statistically dual' to sweeping input in some cases.

The paper shows that all three measures grow \textit{exponentially} with the depth of the network, but not with the width.

\paragraph{Connection to Trajectory Length}
In fact, this is due to an underlying connection of all three measures to another quantity, \textit{trajectory length} -- how a 1-D curve in input space changes in length as it propagates through the network. It is proved \cite{raghu2016expressivity} that the trajectory length of an input grows exponentially in the depth of a network but not the width: 

\begin{theorem}
\emph{Bound on Growth of Trajectory Length}
\label{thm_lb_perturbation_bias}
Let $F_W$ be a  hard tanh random neural network and $x(t)$ a one dimensional trajectory in input space. Define $z^{(d)}(x(t)) = z^{(d)}(t)$ to be the image of the trajectory in layer $d$ of $F_W$, and let $l(z^{(d)}(t)) = \int_t \norm{\frac{d z^{(d)}(t)}{ d t} }dt$ be the arc length of $z^{(d)}(t)$. Then
\[   \mbold{E}\left[l(z^{(d)}(t))\right] \geq  O\left( \left( \frac{\sigma_w}{(\sigma_w^2 + \sigma_b^2)^{1/4}} \cdot \frac{\sqrt{ k}}{\sqrt{ \sqrt{\sigma_w^2 + \sigma_b^2} + k}} \right)^d \right) l(x(t)) \]
\end{theorem}

\begin{figure}
\centering
\begin{tabular}{cc}
(a)\adjincludegraphics[width=0.4\linewidth]{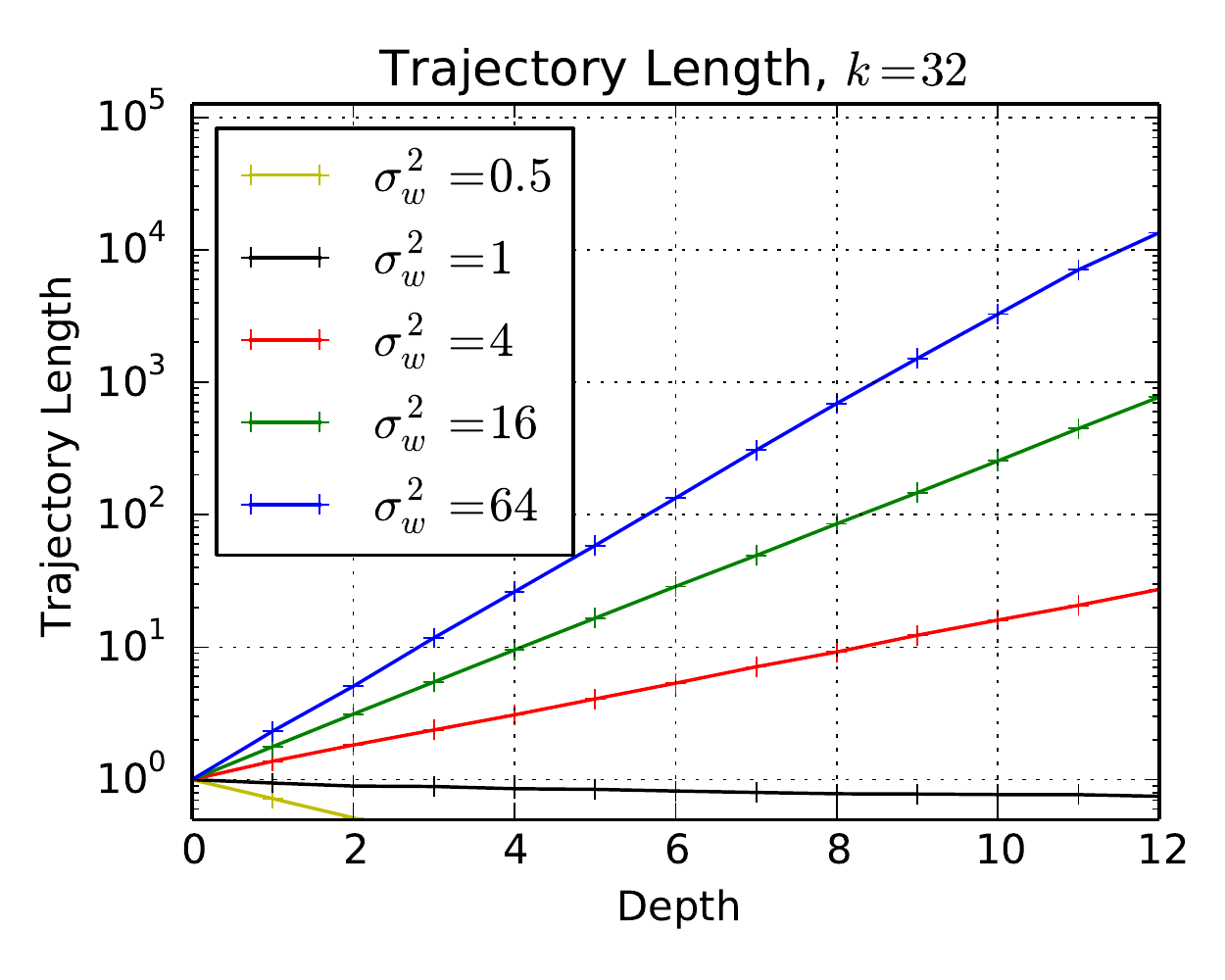} &
(b)\adjincludegraphics[width=0.4\linewidth]{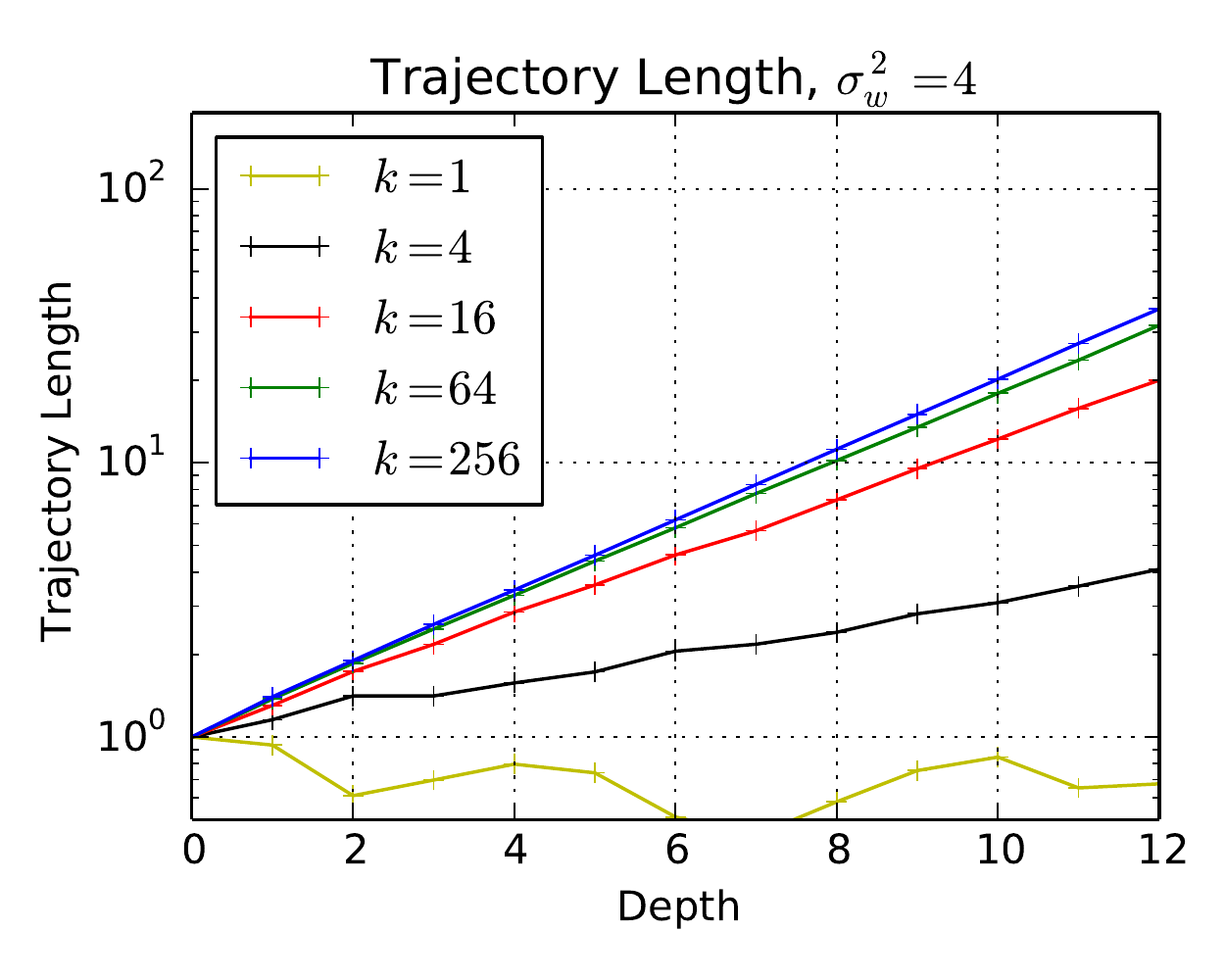} \\
(c)\adjincludegraphics[width=0.4\linewidth]{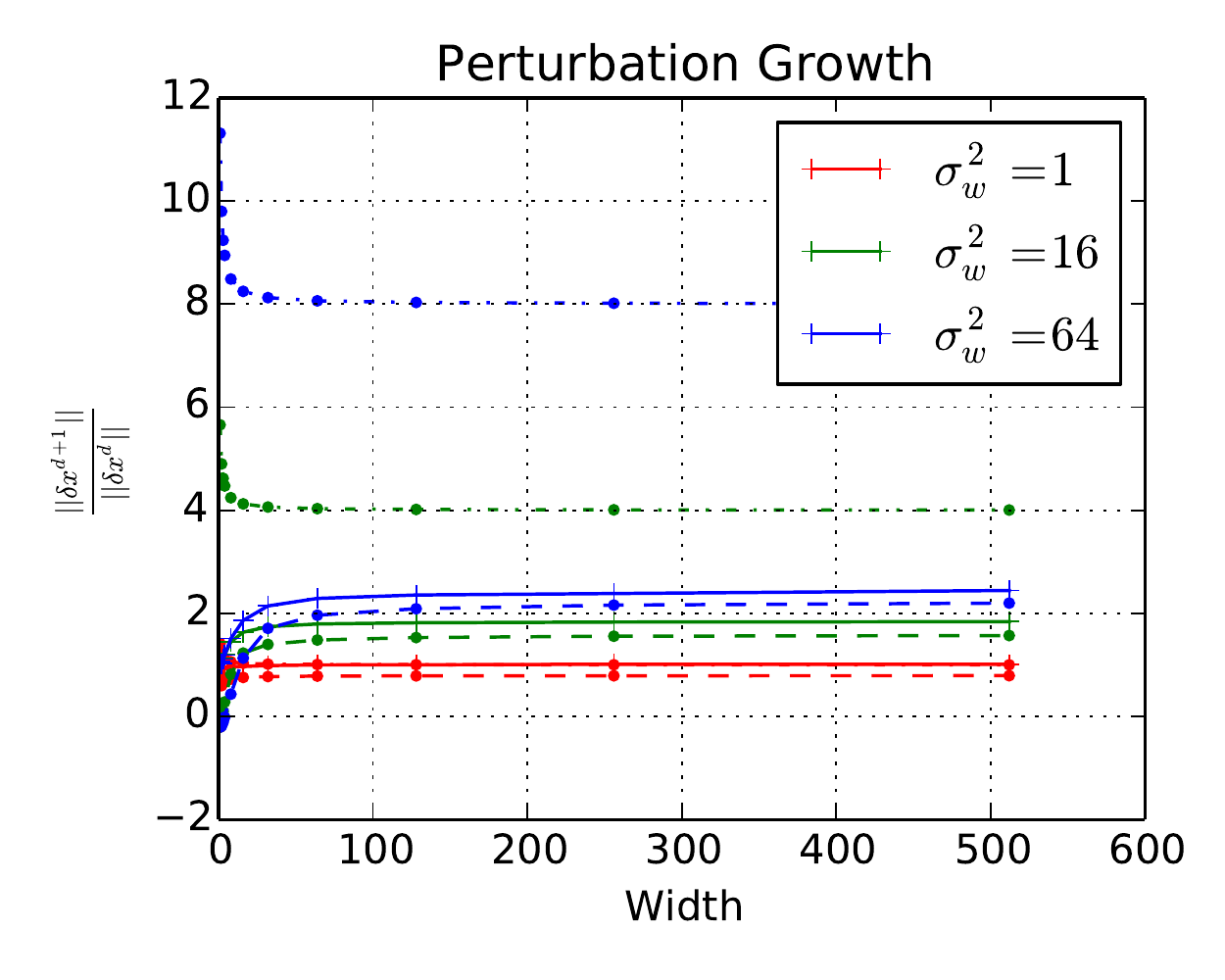} &
(d)\adjincludegraphics[width=0.4\linewidth]{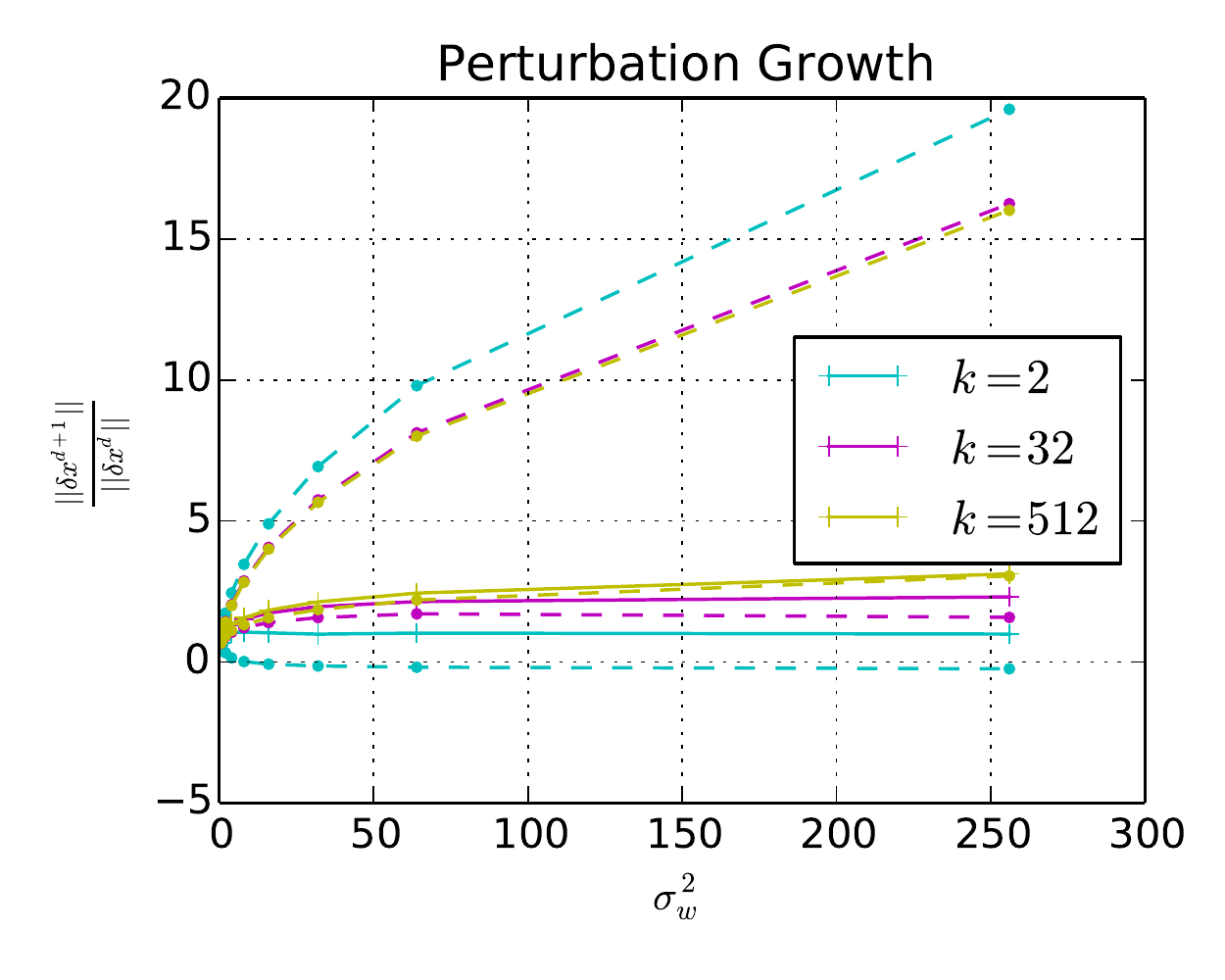} 
\end{tabular}
\caption{
The exponential growth of trajectory length with depth, in a random deep network with hard-tanh nonlinearities.
A circular trajectory is chosen between two random vectors.
The image of that trajectory is taken at each layer of the network, and its length measured.
\emph{(a,b)}
The trajectory length vs. layer, in terms of the network width $k$ and weight variance $\sigma_w^2$, both of which determine its growth rate.
\emph{(c,d)}
The average ratio of a trajectory's length in layer $d+1$ relative to its length in layer $d$.
The solid line shows simulated data, while the dashed lines show upper and lower bounds (Theorem \ref{thm_lb_perturbation_bias}).
Growth rate is a function of layer width $k$, and weight variance $\sigma^2_w$.
\label{fig growth}
}
\end{figure}

This is also verified empirically (Figure \ref{fig growth}).

Theoretical intuition is the provided for the \textit{direct proportionality} of transitions, activation patterns and dichotomies to trajectory length, and is further confirmed through experiments (\cite{raghu2016expressivity}).

\begin{figure}
\centering
\adjincludegraphics[width=1.07\linewidth]{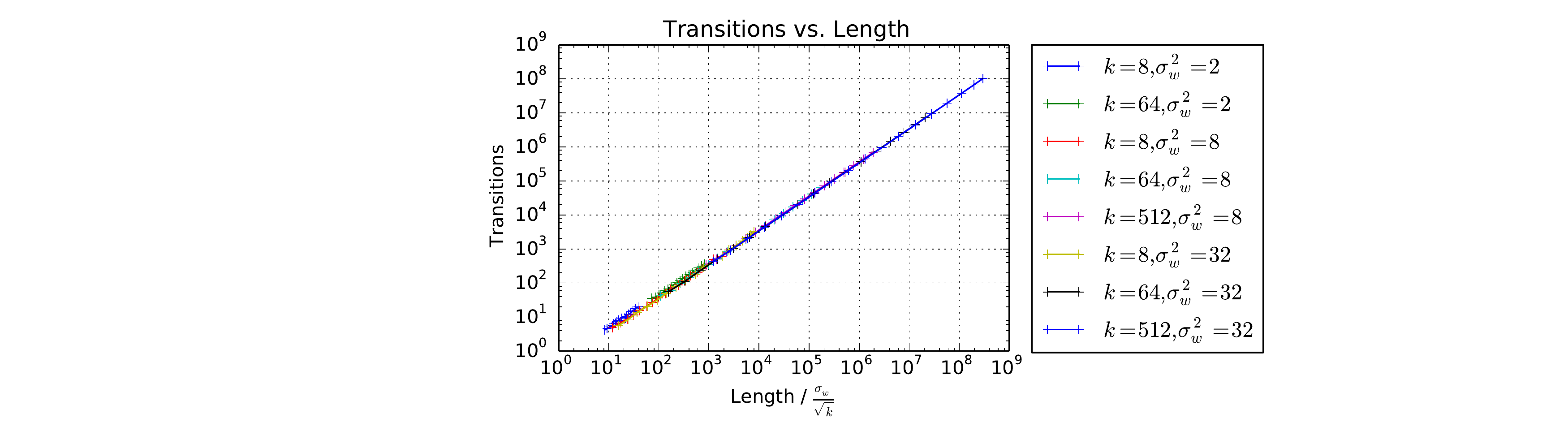}
\caption{
The number of transitions is linear in trajectory length.
Here we compare the empirical number of sign changes to the length of the trajectory,
for images of the same trajectory at different layers of a hard-tanh network. 
We repeat this comparison for a variety of network architectures, with different network width $k$ 
and weight variance $\sigma^2_w$.
\label{fig linear transitions}
}
\end{figure}

\section{The effect of Training: Trading Off Expressivity and Stability}
\label{sec_training}

The paper then (\cite{raghu2016expressivity}) explores the effect of training on the measures of expressivity. Most importantly, note that an exponential depth dependence, as demonstrated at the start of training, makes the resulting function very sensitive to perturbations, not a desired feature in a trained network. 

\begin{figure}
\centering
\adjincludegraphics[width=0.75\linewidth]{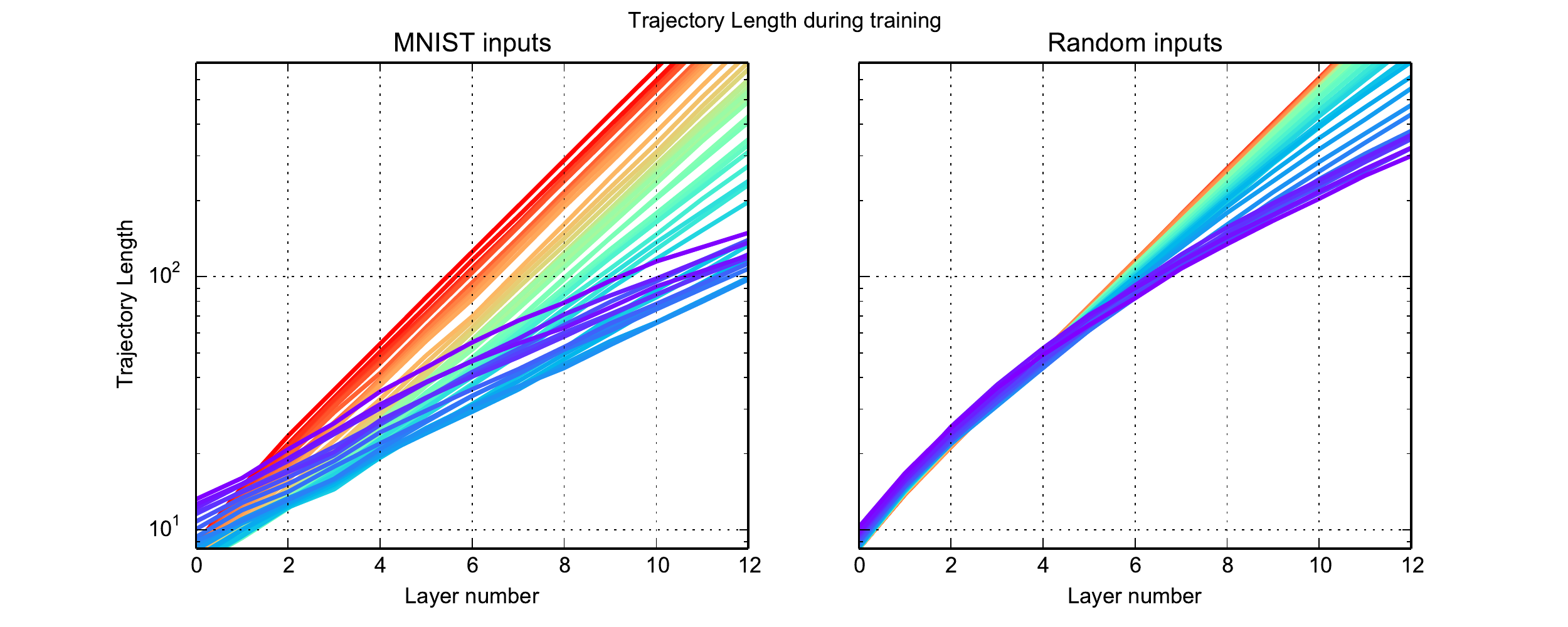}
\caption{
Training acts to stabilize the input-output map by decreasing trajectory length for $\sigma_w$ large. 
The left pane plots the growth of trajectory length as a circular interpolation between two MNIST datapoints is propagated through the network, at different train steps. Red indicates the start of training, with purple the end of training. Interestingly, and supporting the observation on remaining depth, the first layer appears to increase trajectory length, in contrast with all later layers,
suggesting it is being primarily used to fit the data. The right pane shows an identical plot but for an interpolation between \textit{random points}, which also display decreasing trajectory length, but at a slower rate. Note the output layer is not plotted, due to artificial scaling of length through normalization. The network is initialized with $\sigma_w^2 = 16$. A similar plot is observed for the number of transitions (see Appendix.)
\label{mnist_traj_trans}
}
\end{figure}

When weights are initialized with large $\sigma_w^2$, training increases stability by reducing trajectory length and transitions during the training process (Figure \ref{mnist_traj_trans}).

\begin{figure}
\centering
\adjincludegraphics[width=0.75\linewidth]{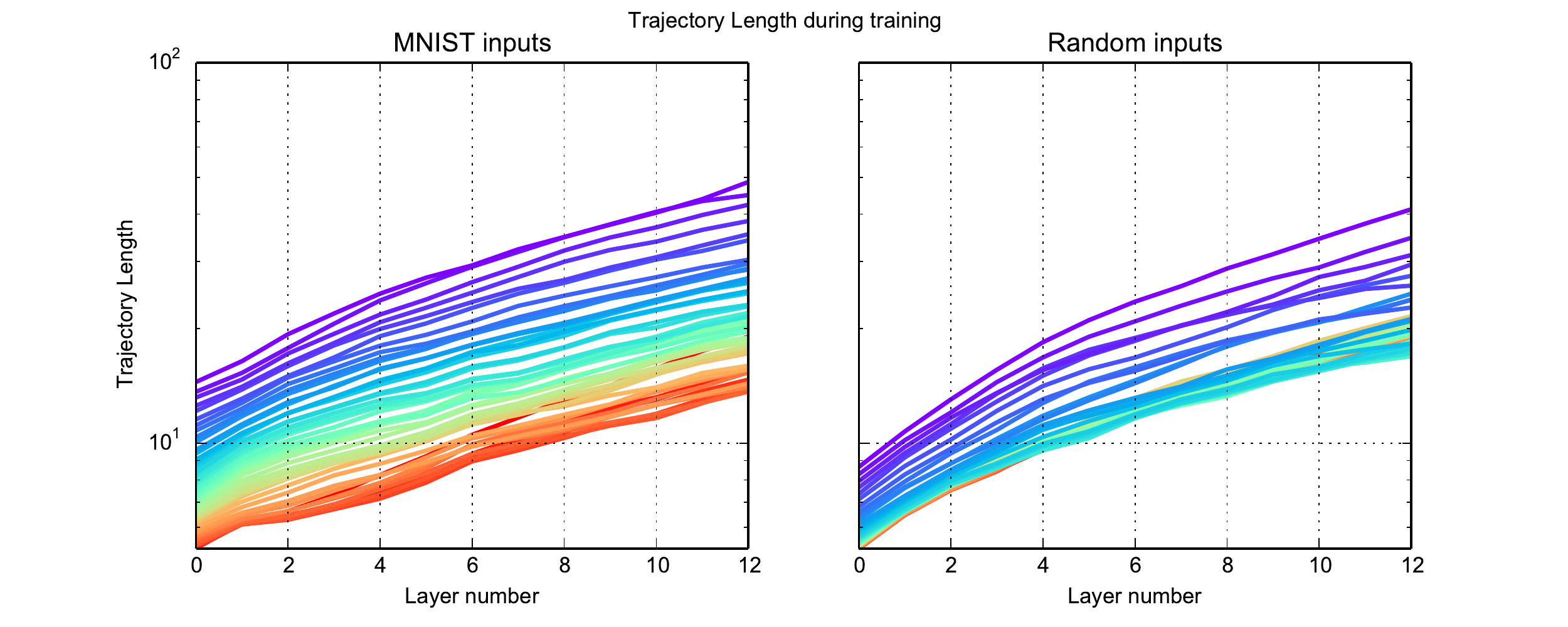}
\caption{
Training increases expressivity of input-output map for $\sigma_w$ small.
The left pane plots the growth of trajectory length as a circular interpolation between two MNIST datapoints is propagated through the network, at different train steps. Red indicates the start of training, with purple the end of training. We see that the training process \textit{increases} trajectory length, likely to increase the expressivity of the input-output map to enable greater accuracy. The right pane shows an identical plot but for an interpolation between \textit{random points}, which also displays increasing trajectory length, but at a slower rate. Note the output layer is not plotted, due to artificial scaling of length through normalization. The network is initialized with $\sigma_w^2 = 3$. 
\label{mnist_traj_trans_sigma_3}
}
\end{figure}

When the network is initialized with too small a $\sigma_w^2$ however, this also has the potential to adversely affect performance as the function at initialization might not offer enough expressiveness to fit the target. In this case, we see that the training process monotonically \textit{increases} the trajectory length and number of transitions (Figure \ref{mnist_traj_trans_sigma_3}.)

In summary, the paper \cite{raghu2016expressivity} concludes that training trades off between achieving enough expressiveness and simultaneously trying to maintain stability.

\section{Trained Networks: Power of Remaining Depth}
\label{sec_trained}
\begin{figure}
\centering
\adjincludegraphics[width=0.75\linewidth]{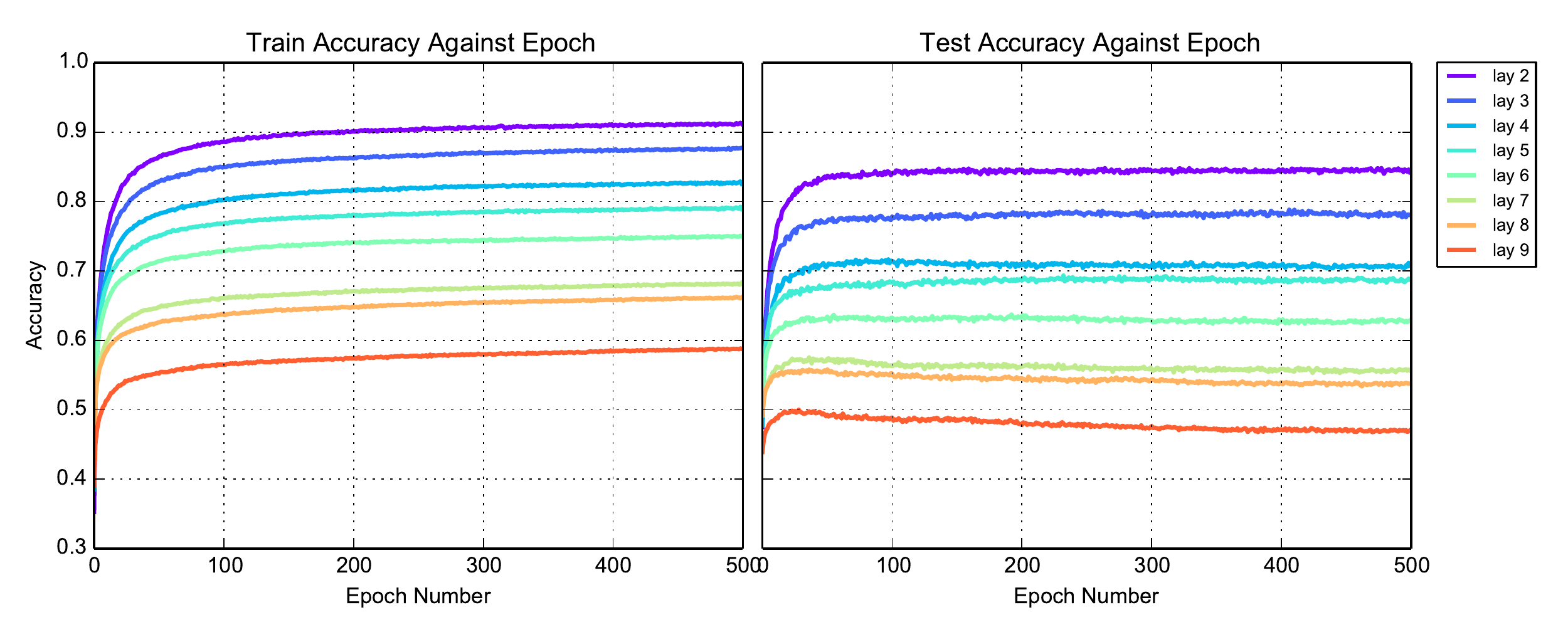}
\caption{
Demonstration of expressive power of remaining depth on MNIST.
Here we plot train and test accuracy achieved by training exactly one layer of a fully connected neural net on MNIST. The different lines are generated by varying the hidden layer chosen to train. All other layers are kept frozen after random initialization.
\label{mnist_ffn}
}
\end{figure}

The expanding trajectory length suggests that the effect of parameter choices earlier in earlier layers is \textit{amplified} by later layers. Combining this with the exponential increase in dichotomies with depth, this suggests that the expressive power of the parameters, and thus layers, is related to the \textit{remaining depth} of the network after that layer. The paper demonstrates this in practice, with experiments on MNIST and CIFAR-10 (Figure \ref{mnist_ffn}).

\small
\setlength{\bibsep}{0pt plus 0.2ex}

\bibliographystyle{unsrtnat}

\bibliography{deep_expressivity}

\begin{thebibliography}{9}
\providecommand{\natexlab}[1]{#1}
\providecommand{\url}[1]{\texttt{#1}}
\expandafter\ifx\csname urlstyle\endcsname\relax
  \providecommand{\doi}[1]{doi: #1}\else
  \providecommand{\doi}{doi: \begingroup \urlstyle{rm}\Url}\fi

\bibitem[{Raghu} et~al.(2016){Raghu}, {Poole}, {Kleinberg}, {Ganguli}, and
  {Sohl-Dickstein}]{raghu2016expressivity}
Maithra {Raghu}, Ben {Poole}, Jon {Kleinberg}, Surya {Ganguli}, and Jascha
  {Sohl-Dickstein}.
\newblock {On the expressive power of deep neural networks}.
\newblock \emph{ArXiv e-prints}, June 2016.
\newblock URL \url{https://arxiv.org/abs/1606.05336}.

\bibitem[Hornik et~al.(1989)Hornik, Stinchcombe, and
  White]{hornik1989multilayer}
Kurt Hornik, Maxwell Stinchcombe, and Halbert White.
\newblock Multilayer feedforward networks are universal approximators.
\newblock \emph{Neural networks}, 2\penalty0 (5):\penalty0 359--366, 1989.

\bibitem[Cybenko(1989)]{cybenko1989approximation}
George Cybenko.
\newblock Approximation by superpositions of a sigmoidal function.
\newblock \emph{Mathematics of control, signals and systems}, 2\penalty0
  (4):\penalty0 303--314, 1989.

\bibitem[Eldan and Shamir(2015)]{eldan2015power}
Ronen Eldan and Ohad Shamir.
\newblock The power of depth for feedforward neural networks.
\newblock \emph{arXiv preprint arXiv:1512.03965}, 2015.

\bibitem[Telgarsky(2015)]{telgarsky2015representation}
Matus Telgarsky.
\newblock Representation benefits of deep feedforward networks.
\newblock \emph{arXiv preprint arXiv:1509.08101}, 2015.

\bibitem[Martens et~al.(2013)Martens, Chattopadhya, Pitassi, and
  Zemel]{martens2013representational}
James Martens, Arkadev Chattopadhya, Toni Pitassi, and Richard Zemel.
\newblock On the representational efficiency of restricted boltzmann machines.
\newblock In \emph{Advances in Neural Information Processing Systems}, pages
  2877--2885, 2013.

\bibitem[Bianchini and Scarselli(2014)]{bianchini2014complexity}
Monica Bianchini and Franco Scarselli.
\newblock On the complexity of neural network classifiers: A comparison between
  shallow and deep architectures.
\newblock \emph{Neural Networks and Learning Systems, IEEE Transactions on},
  25\penalty0 (8):\penalty0 1553--1565, 2014.

\bibitem[Poole et~al.(2016)Poole, Lahiri, Raghu, Sohl-Dickstein, and
  Ganguli]{poole2016exponential}
Ben Poole, Subhaneil Lahiri, Maithra Raghu, Jascha Sohl-Dickstein, and Surya
  Ganguli.
\newblock Exponential expressivity in deep neural networks through transient
  chaos.
\newblock In \emph{Advances In Neural Information Processing Systems}, pages
  3360--3368, 2016.

\bibitem[Pascanu et~al.(2013)Pascanu, Montufar, and Bengio]{pascanu2013number}
Razvan Pascanu, Guido Montufar, and Yoshua Bengio.
\newblock On the number of response regions of deep feed forward networks with
  piece-wise linear activations.
\newblock \emph{arXiv preprint arXiv:1312.6098}, 2013.

\end{thebibliography}

\clearpage

\end{document}